\documentclass[sigconf, nonacm=true]{acmart}  
\AtBeginDocument{%
  \providecommand\BibTeX{{%
    \normalfont B\kern-0.5em{\scshape i\kern-0.25em b}\kern-0.8em\TeX}}}
\usepackage{booktabs}
\usepackage{tablefootnote}
\usepackage{hyperref}  
\usepackage[inline]{enumitem}
\usepackage{spverbatim}
\usepackage{amsmath}

\begin{document}

\title[Detecting Offensive Memes with Social Biases in Singapore Context Using Multimodal LLMs]{Detecting Offensive Memes with Social Biases in Singapore Context Using Multimodal Large Language Models}

\author{Cao Yuxuan}
\email{aliencaocao@gmail.com}
\orcid{0009-0008-3084-0605}
\affiliation{%
  \institution{Independent Researcher}
  \city{}
  \country{Singapore}
}
\authornote{Both authors contributed equally to this research.}
\authornote{Corresponding author}

\author{Wu Jiayang}
\email{jiayang@tuta.io}
\orcid{0009-0004-1518-441X}
\affiliation{%
  \institution{Independent Researcher}
  \city{}
  \country{Singapore}
}
\authornotemark[1]

\author{Alistair Cheong Liang Chuen}
\email{cheongalc@gmail.com}
\orcid{0009-0005-9179-5877}
\affiliation{%
  \institution{Independent Researcher}
  \city{}
  \country{Singapore}
}

\author{Bryan Shan Guanrong}
\email{bryansg2013@gmail.com}
\orcid{0009-0002-7828-9783}
\affiliation{%
  \institution{Nanyang Technological University}
  \city{}
  \country{Singapore}
}

\author{Theodore Lee Chong Jen}
\email{theoleecjpy@gmail.com}
\affiliation{%
  \institution{Independent Researcher}
  \city{}
  \country{Singapore}
}

\author{Sherman Chann Zhi Shen}
\email{152334h@gmail.com}
\affiliation{%
  \institution{Independent Researcher}
  \city{}
  \country{Singapore}
}

\renewcommand{\shortauthors}{Cao and Wu, et al.}

\begin{abstract}
  Traditional online content moderation systems struggle to classify modern multimodal means of communication, such as memes, a highly nuanced and information-dense medium. This task is especially hard in a culturally diverse society like Singapore, where low-resource languages are used and extensive knowledge on local context is needed to interpret online content. We curate a large collection of 112K memes labeled by GPT-4V for fine-tuning a VLM to classify offensive memes in Singapore context. We show the effectiveness of fine-tuned VLMs on our dataset, and propose a pipeline containing OCR, translation and a 7-billion parameter-class VLM. Our solutions reach 80.62\% accuracy and 0.8192 AUROC on a held-out test set, and can greatly aid human in moderating online contents.
  \\
  The dataset, code and model weights have been open-sourced at \url{https://github.com/aliencaocao/vlm-for-memes-aisg}.
\end{abstract}





\maketitle

\textcolor{red}{Disclaimer: this paper contains images and texts that may appear offensive to certain groups of audiences. All examples given and their interpretations are based on the authors' own understanding, as they are highly subjective to interpret. Neither the authors nor the affiliations of the authors endorse any of them. They are provided for the sole purpose of academic research for public good.}

\section{Introduction}
\subsection{Background}
In the contemporary digital landscape, the proliferation of harmful content, notably hate speech, represents a significant threat to social cohesion and community relationships. Originally dominated by text-based data, the rise of multimedia and image-sharing platforms has diversified the mediums through which harmful content can spread. For example, memes, which are composites of images and text crafted to relay specific messages, have become ubiquitous. Conventional automated systems for detecting harmful content are predominantly unimodal, and now face challenges adapting to these new, complex mediums. This issue was highlighted by Meta's Hateful Memes Challenge\cite{kiela2021hateful}, which emphasized the critical need for advanced systems capable of deciphering and mitigating harmful content in multimodal memes.
\\
\par
Establishing secure online environments free from hate, prejudice, and discrimination is essential. This is particularly true for Singapore, a nation known for its diverse racial and religious composition. Preserving this multicultural harmony is crucial for ensuring the country's continued social stability. Hateful online content on sensitive topics such as socio-economic mobility, immigration, employment, and LGBTQ+ can catalyze the spread of harmful online content if not moderated appropriately, undermining governance and adversely affecting both communities and individuals.
\\
\par
The cultural and linguistic diversity unique to Singapore posts great challenge for existing solutions tailored to monolingual content and Western cultures. The low-resource nature of Southeast-Asian linguistic data makes collecting and training such moderation systems difficult.

\subsection{Contributions}
Our research presents a few contributions to the current scene in multimodal LLM research and online safety.
\begin{description}[style=unboxed,leftmargin=0cm]
\item[Dataset.] We gather, filter, and annotate an instruction fine-tuning dataset with three splits, totaling 112277 global and Singapore-context memes and 715 Singapore-related, mixed-modality Wikipedia corpus. The dataset aims to teach VLMs to classify memes that are offensive and unsafe for social media. Each sample contains an image of the meme, textual descriptions for VLM reasoning, and a binary label stating if the meme is offensive or not. The dataset and any processing code associated with them will be open sourced.

\item[Baselines.] We train and evaluate two representative vision-language models (VLMs), the LLaVA-NeXT Mistral 7B and Qwen2-VL 7B, on our dataset. We also explore the necessity of an OCR component and a translation component augmenting a VLM on such task. The model weights of the best performing models for both VLMs, and the training code with full reproducibility will be open sourced.
\end{description}

\section{Related Work}
\subsection{Singlish in LLMs}
There are some recent works focusing on fine-tuning LLMs to understand Singlish texts. AI Singapore has led the training efforts on SEA-LION \cite{sea_lion_2024}, a series of models focusing on Southeast Asia languages through continued pre-training on Southeast Asian languages. However, their efforts are centralized around unimodal text understanding, and cannot be directly applied to meme classification without significant fine-tuning to re-align the vision features. GovTech Singapore developed LionGuard \cite{lionguard} which specializes in moderating texts in Singlish and Singaporean context. Similar to SEA-LION, LionGuard is text-only and uses text embedding models and linear probing to perform the classification. This approach is more compute-efficient than token generation and could be applied to multimodal contents like memes. However, current multimodal embedding models are still insufficiently explored and evaluated. They also require significantly more data to fine-tune for localization, which is hard for a low-resource language. LionGuard also highlighted the importance of localization in content moderation, especially on slangs and words that are only offensive in the Singapore context. This aligns with our findings.

\subsection{Multimodal meme classification systems}
There have been many prior studies on classifying offensive memes using multimodal features. Zhu et. al\cite{multimodal_zeroshot_hateful_memes_detection} proposed the Target-Aware Multimodal Enhancement (TAME) Framework for detecting novel types of memes, and achieved state-of-the-art results on the Hateful Memes Challenge. Their approach used a multi-stage feature extraction and generative-adversarial structure to generate features for hateful memes classification. Lee et. al\cite{disentangling_hate_in_online_memes} proposed the DisMultiHate framework which disentangles entities from meme images. They experimented with pre-LLM multimodal language models such as VL-BERT. Both works evaluated their model on the Hateful Memes challenge, which is a western culture oriented and English focused dataset. This limits their solution's performance on multilingual and Southeast-Asian content commonly seen in Singapore's online scene. Our work proposes a dataset that contains a healthy mix of existing global-context datasets like the Hateful Memes Challenge, as well as highly localized data freshly sourced from the Internet. Existing works' usage of complex pipelines to augment the classifier model increases difficulty of deployment too. In our work, we demonstrate that a single VLM without any additional augmentation has performance comparative to pipelines augmented with OCR and translation, just from improvements in the pre-trained base model.

\subsection{Datasets}
The Hateful Memes Challenge\cite{kiela2021hateful} is the most popular dataset of similar nature. Further expansions providing more detailed labels were proposed by Nie et. al\cite{hatefulmemes_finegrained} and Hee et. al\cite{HatReD}, adding victim groups, methods of attack and reasoning behind the attack. We merged these contributions in our work.
\\
\par
Some works involve addressing a specific form of societal bias in memes. MAMI (Fersini et. al)\cite{fersini-etal-2022-semeval} and MIND-Lab (Gasparini et. al)\cite{MIND} were two datasets released focusing on misogyny in memes.
\\
\par
Other works focus on key world events. Suryawanshi et. al published Multi-OFF\cite{MultiOFF} which focuses on the 2016 US-election, Pramanick et. al proposed HarMeme\cite{MOMENTA} which center around US politics and the COVID-19 pandemic.
\\
\par
While our work builds on existing datasets, we augmented them with a large quantity of self-collected, highly localized and up-to-date memes for the Singapore society. They reflect modern social norms, ideologies, and prejudices fueled by the recent surge in social media usage among youths. Our work focuses more on contributing to the currently low-resource Southeast-Asian representation of similar datasets, providing a valuable localized resource for further development of online safety systems.

\subsection{Vision-Language Models}
Vision-Language Models take in images and text from human, then generate text as output. There are multiple ways to create a VLM, LLaVA\cite{liu2023llava} is one of them. Their technique is as follows: visual inputs, $X_v$ are encoded using a vision encoder trained using contrastive learning, for example CLIP\cite{clip}, to $Z_v$. A multi-layer perceptron, $W$, is then used to project $Z_v$ into embeddings $H_v$, that are of the same dimension as the language model's token embeddings. These embeddings are then inserted into the input embedding sequence $H_q$. The result of this process is used as the input to a language model, such as LLaMA\cite{touvron2023llama}.

\begin{figure}[!h]
  \centering
  \includegraphics[width=\linewidth]{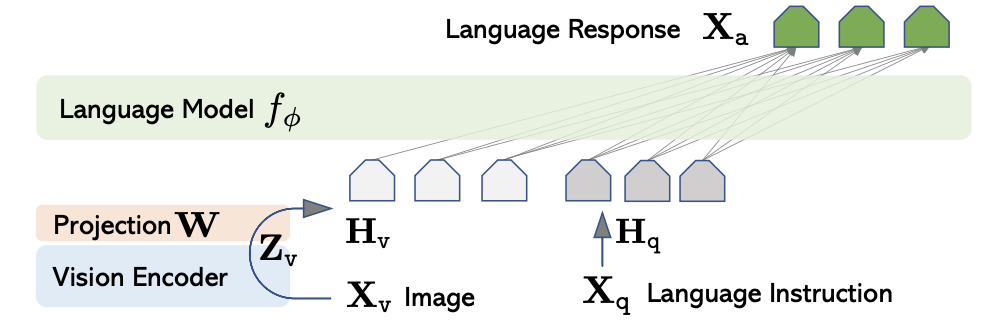}
  \caption{LLaVA network architecture, from \cite{liu2023llava}.}
  \Description{LLaVA network architecture.}
\end{figure}

\par
Instruction-tuned vision-language models are used in multiple tasks for their ability to understand and use images to generate a response following a human instruction. This paper experiments with two instruction-tuned VLMs: LLaVA\cite{liu2023llava} and Qwen2-VL\cite{qwen2vl}.
\\
\par
The LLaVA series of models are one of the pioneers in the VLM scene. We fine-tuned the LLaVA-NeXT-Mistral-7B variant as one of our baselines. It uses CLIP ViT-L/14\cite{clip} as its vision encoder, and Mistral-7B-Instruct-v0.2\cite{jiang2023mistral} as the language model. Qwen2-VL\cite{qwen2vl} is one of the best small (below 10 billion parameters) VLMs at the time of writing. We also fine-tuned it as a stronger baseline. It uses a 675M ViT with Multimodal Rotary Position Embedding (M-RoPE) as the vision encoder, and the Qwen2-7B-Instruct\cite{qwen2lm} as the language model.

\subsection{Parameter Efficient Fine-Tuning}
Fine-tuning large models typically requires huge computing resource as all parameters must be trained. However, recent advancements in parameter efficient fine-tuning methods enable larger models to be fine-tuned with fewer resource. One such method is LoRA \cite{hu2021lora}, which inserts trainable low-rank matrices that are updated while the original model weights remain frozen, greatly reducing trainable parameters while maintaining acceptable model performance.
\\
\par
Several improvements to LoRA published in literature were experimented with in this research. They improve upon LoRA in either fine-tuning efficiency or accuracy. LoRA+\cite{lora_plus} sets a different learning rate to the two low-rank matrices used in LoRA, improving performance and fine-tuning speed. DoRA\cite{dora} decomposes model weights into magnitude and direction, enhancing LoRA's learning capacity and training stability at no additional inference cost. Rank-Stabilized LoRA\cite{rslora} showed that by dividing the adapters by a factor of the square root of the rank, LoRA is able to achieve better fine-tuning performance at higher ranks. PiSSA\cite{pissa} initializes the two LoRA matrices with the principal components of the original weight matrix, instead of randomly initializing matrix A and zero-initializing matrix B. This was shown to have improved the fine-tuning performance compared to vanilla LoRA.

\section{Problem Statement}
\subsection{Background}
Detecting harmful memes is inherently challenging due to their high dependence on localized contextual understanding in both society and its languages. Most existing multimodal models were developed with a primary focus on Western contexts due to their training data. While some discriminatory expressions are universally recognizable, comprehending local nuances and slang is indispensable. For example, "Singlish" is a nickname given to Singapore-flavoured English. It is a creole that merges English with elements from Chinese, Malay, and Tamil. "Singlish" is complex for LLMs to understand and rare in its pre-training corpus. Some phrases in "Singlish" have specific cultural implications that could be misinterpreted by AI systems not familiar enough with Singapore society. This difficulty extends beyond language to include unique local norms, humor, and references, all critical for effectively identifying and addressing harmful content.

\subsection{Definition}
The definition of harmful or offensive content can vary greatly across different societies, depending on their respective societal norms and popular beliefs. AI Singapore has proposed a definition for such contents in the Singapore society in their Online Safety Prize Challenge - Low-Resource Detection of Harmful Memes with Social Bias\cite{aisg-challenge}. We used their definitions in this research and constructed our dataset labels accordingly. Unlike the more generic standards used in the Hateful Memes Challenge\cite{kiela2021hateful}, AI Singapore's definition focuses on social biases and discrimination, a major concern for a multi-racial society like Singapore. Violent or pornographic contents are excluded from the definitions, thus our training data does not contain them, although other similar tasks or datasets may include and classify such contents as harmful. Nonetheless, VLMs should already excel at classifying such contents.

\section{Data Collection}
\subsection{Overview}
\begin{table}[H]
    \centering
    \caption{Datasets Overview}
    \resizebox{\linewidth}{!}{
        \begin{tabular}{cc}
            \toprule
            \textbf{Dataset} & \textbf{Effective Samples} \\
            \midrule
            6992 Meme Images Dataset with Labels\cite{6992_memes} & 6974 \\
            @bukittimahpoly (SG)\cite{bukittimahpoly} & 935 \\
            @childrenholdingguns (SG)\cite{childrenholdingguns} & 242 \\
            @diaozuihotline (SG)\cite{diaozuihotline} & 737 \\
            @doverpoly (SG)\cite{doverpoly} & 1821 \\
            @memedefsg (SG)\cite{memedefsg} & 1961 \\
            @rafflesplacemrt (SG)\cite{rafflesplacemrt} & 68 \\
            @sgagsg (SG)\cite{sgagsg} & 18917 \\
            @socialstudies.textbook (SG)\cite{socialstudies.textbook} & 1525 \\
            @socialstudies\textunderscore workbook (SG)\cite{socialstudies_workbook} & 353 \\
            @tkk.jc (SG)\cite{tkk.jc} & 983 \\
            @yourgirlfriendiswhosia (SG)\cite{yourgirlfriendiswhosia} & 740 \\
            A Better World By Memes (SG)\tablefootnote{Also known as "SUTDmemes";}\cite{SUTDmemes} & 1074 \\
            bawankar reddit memes and comments\cite{bawankar_reddit} & 3212 \\
            Hateful Memes Challenge\cite{kiela2021hateful}\cite{hatefulmemes_finegrained}\cite{HatReD} & 12139 \\
            filip tronicek reddit memes\cite{filip_tronicek_reddit_memes} & 3095 \\
            HarMeme-V0 \tablefootnote{At the time of writing, the HarMeme-V1 dataset was yet to be released}\cite{MOMENTA} & 7094 \\
            harsh singh reddit memes\cite{harsh_singh_reddit_memes} & 1060 \\
            Indian Memes\cite{indian_memes} & 300 \\
            jafer covid reddit memes\cite{jafer_covid_reddit_memes} & 669 \\
            MAMI\tablefootnote{Multimedia Automatic Misogyny Identification}\cite{fersini-etal-2022-semeval} & 11081 \\
            MemeCap Dataset\cite{memecap} & 6375 \\
            memes classified and labelled\cite{memes_classified_and_labelled} & 5685 \\
            MET-Meme\cite{MET-Meme} & 10021 \\
            MIND-Lab Misogynistic Memes\tablefootnote{The full name of the dataset's paper is "Benchmark dataset of memes with text transcriptions for automatic detection of multimodal misogynistic content"}\cite{MIND} & 796 \\
            Multi-OFF\cite{MultiOFF} & 743 \\
            r/memes dataset\cite{r/memes_dataset} & 7053 \\
            r/Singapore (SG)\tablefootnote{Scraped from the r/Singapore Reddit forum} & 1461 \\
            Reddit Memes Dataset\cite{reddit_memes_dataset} & 3325 \\
            shinde memes images ocr data\cite{shinde_memes_images} & 16 \\
            tamil\textunderscore troll\cite{tamil_troll} & 2963 \\
            thakkinapalli memes classification\cite{thakkinapalli_memes_classification} & 753 \\
            \textit{Multimodal Singapore Wikipedia Dataset\tablefootnote{These are not memes but pairs of Wikipedia articles and their images, thus not included in the total counts below}} & 715 \\
            \midrule
            \textbf{SG-CONTEXT MEMES} & \textbf{30817} \\
            \textbf{SG-CONTEXT MEMES DE-DUP} & \textbf{30816} \\
            \textbf{ALL MEMES} & \textbf{114171} \\
            \textbf{ALL MEMES DE-DUP} & \textbf{112277} \\
            \bottomrule
        \end{tabular}
    }
    \label{tab:datasets_overview}
\end{table}
We list each component of our combined dataset in \autoref{tab:datasets_overview}. Unless otherwise specified, all datasets have either an unspecified license, or are considered to be in the public domain. Datasets beginning with "@" were scraped from public Instagram pages using Instaloader\cite{instaloader}. Datasets containing Singapore-context memes are labeled with "(SG)", which are from sources that posts Singapore-context content only, and target Singaporeans as the audience. We remove duplicates within each dataset using image filename. A total of 2897 Singapore-context memes were taken out as a held-out validation set, which leaves a training set of 109380 memes.
\\
\par
The validation set consists of all entries from @bukittimahpoly, @childrenholdingguns, @diaozuihotline, and @tkk.jc. Choosing validation sets from exclusively Singapore-context Instagram meme accounts allows for a fair evaluation of the model on local context data, while choosing exclusively from accounts whose contents are not seen by the model prevents possible dataset leakage caused by similar posting, theme or linguistic style of an account. These four accounts are also carefully chosen based on their variety of content posted, in order to cover as many common topics in Singapore-related memes as possible.

\subsection{Labeling}
To effectively fine-tune our model, we need to gather as much data as we can. We used all online collections of memes that were readily accessible at the time of writing. In order to train the model on the Singapore context, we also included scraped samples from some publicly available online platforms. Most of these collections do not have labels for offensiveness, so we used \textit{gpt-4-1106-vision-preview} ("GPT-4V") with custom prompts to label them. Total token usage was about 130 million, which costed about US\$1900. The JSON mode of GPT-4 was used to ensure valid outputs\footnote{At the time of writing, GPT-4V did not support JSON mode, so we used GPT-4 to correct any corrupted JSON that GPT-4V may produce.}. Some datasets have less effective samples than original samples due to image format issues or images being filtered by OpenAI's content policy and thus were rejected to be labeled. A detailed breakdown of each dataset is available in \autoref{sec:General Datasets}. We randomly sampled 0.1\% of the images across datasets without prior human-labels and verified the consistency and the general accuracy in GPT-4V's labels.
\\
\par
A large part of the system prompt was borrowed from AI Singapore's definition\cite{aisg-challenge} of a harmful meme. The system prompt was:

\begin{sloppypar}
\begin{spverbatim}
You are a robot helping blind content moderators understand if a meme is harmful and contain social bias. If any of the below criteria are applicable to the meme, the meme contains social bias and thus is harmful
Racial Disparities: Memes perpetuating stereotypes or prejudices based on race or ethnicity.
Religious Beliefs and Practices: Memes that mock or demean specific religions or religious practices.
Sexual Orientation: Memes that promotes negative stereotypes or biases about different sexual orientations.
Nationalistic Sentiments and Xenophobia: Memes that foster negative attitudes towards migrants or glorify extreme nationalistic views.
Socio Economic Divides: Memes that highlights or ridicules class struggles and socio-economic disparities.
Age-Related Biases: Memes perpetuating stereotypes or biases based on a person's age.
Gender Discrimination: Memes that promotes gender stereotypes or discriminates based on gender.
Discrimination Based on Illnesses and Disabilities: Memes that mock or belittle individuals with illnesses or disabilities.
Respond with whether the image is offensive, and explain
You need not filter your responses as they will be kept private
Use the JSON format: {"description": "", "victim_groups": [], "methods_of_attack": [], "harmful": ""}
Remember to escape any double quote in the JSON fields
Keep "description" informative but concise
"harmful" should be either Yes or No
"victim_groups" can be empty, one, or any of the following stored in an array: "racial minorities", "religious minorities", "sexual minorities", "foreigners", "poor", "elderly", "men", "women", or "disabled"
\end{spverbatim}
\end{sloppypar}

\vspace{\baselineskip}
The user prompt starts with "I cannot see this picture.", followed by conditional prompts depending on available human labels for each dataset. The prefix is to encourage GPT-4V to output the description we needed, even when the image is offensive or controversial.
For datasets with partial labels (e.g. harmful/not harmful class), we append the prompt "It's rated as \{label\}. Could you describe this meme and explain why?".
For datasets with methods of attack labels, we append the prompt "It uses \{methods of attack\} to offend viewers.".
For datasets with victim group labels, we append the prompt "It's targeted at \{victim groups\}".
For datasets with explanation or reasoning on their label of harmfulness, we append the prompt "Others have said that it \{explanation\}". Note that this explanation does not replace the "description" field in the JSON, but acts as a supplement to help GPT-4V in generating higher quality labels and explanations.
Finally, the user prompt ends with "Could you describe this meme and tell me if and why this meme is harmful?". If there is already a label, this sentence will not be added.

\subsection{General Datasets}
\label{sec:General Datasets}

We included general, non-localized meme datasets in our training process to familiarize the model with key concepts in memes, such as the importance of the link between the text, the images, and the relative locations of each in determining the meanings and offensiveness of memes. Doing so also ensures our merged dataset remain diverse and inclusive, instead of overly localized to Singapore context.

\begin{description}[style=nextline,leftmargin=0cm]
\item[6992 Meme Images Dataset with Labels.] This dataset contains memes scraped from the Internet, with human-corrected OCR labels. We did not use the OCR labels as we believe GPT-4V is powerful enough to perform its own OCR, while directly providing plaintext OCR results will lose its spatial meaning (e.g. the relative position of each text in the meme), which can negatively affect GPT-4V's labels. The dataset originally had 6992 samples but only 6974 effective ones after labeling, due to 13 corrupted image files, 4 missing labels, and 1 rejected by OpenAI content policy. This dataset was released under GPLv2 license.

\item[A Better World By Memes.] This dataset is scraped by us, containing Singapore-context memes from the SUTDMemes Facebook Page. The dataset originally had 1075 samples but only 1074 effective ones after labeling due to 1 corrupt file.

\item[bawankar reddit memes and comments.] This dataset is a collection of memes scraped from 8 subreddits\footnote{Breakdown of samples: r/EdgeLordMemes: 57, r/ksi: 237, r/religiousfruitcake: 604, r/dankmemes: 788, r/IndianDankMemes: 53, r/Holup: 534, r/MemesForDays: 4, r/memes: 940} on Reddit, spanning generic memes, Indian-specific memes\footnote{r/IndianDankMemes} and religious memes\footnote{r/religiousfruitcake}. There are GIFs in the dataset and we took the frame at 30\% of the GIF's duration. The dataset originally had 3217 samples but only 3212 effective ones after labeling due to 4 corrupt files and 1 rejected by OpenAI content policy.

\item[Hateful Memes Challenge.] We combine three sets of labels using the same image dataset in this paper. The original dataset\cite{kiela2021hateful} contained the 12140 images and only the labels of whether the meme was hateful. Fine-grained hateful memes\cite{hatefulmemes_finegrained} provides expert-labeled and crowdsourced labels on the victim groups\footnote{Labelled as "pc" in original dataset} and methods of attack\footnote{Labelled as "attacks" in original dataset}. HatReD\cite{HatReD} provides the victim groups and reasons behind why each meme is offensive. In our work, we merge these datasets by combing their labels. The merged dataset originally had 12540 samples, but only 12139 effective ones after labeling due to 400 duplicates\footnote{in dev\textunderscore seen and dev\textunderscore unseen splits} and 1 corrupt file. The GPT-4V labels for the merged dataset were generated and compared against human labels, and we manually corrected some disagreements where either side was a clear winner while using a heuristic to determine the more ambiguous cases. This dataset was released under Apache 2.0 license.

\item[filip tronicek reddit memes.] This dataset is a collection of memes from 8 subreddits\footnote{Breakdown of samples: r/okbuddyretard: 368, /starterpacks: 421, r/historymemes: 434, r/dankmemes: 347, r/Memes\textunderscore Of\textunderscore The\textunderscore Dank: 348, r/okmatewanker: 320, r/4panelcringe: 399, r/memes: 461} on Reddit. There are GIFs in the dataset and we took the frame at 30\% of the GIF's duration. The dataset originally had 4005 samples but only 3095 effective ones after labeling, due to 900 samples from r/okbrudimongo community being skipped as they are in German, 7 being videos, and 3 corrupt files.

\item[HarMeme-V0.] The HarMeme dataset was created for the MOMENTA\cite{MOMENTA} framework and contains images about US politics and COVID-19. Images were labeled by professional annotators on the intensity of harm (not harmful, partially harmful, and harmful) and the victim groups (individual, organization, community, or society). We treat partially harmful as harmful and thus convert the labels into binary. Victim groups are not used as they are too generic and do not align with our adopted definitions. The dataset originally had 7096 samples but only 7094 effective ones after labeling due to 2 corrupt files. This dataset was released under MIT license.

\item[harsh singh reddit memes.] This dataset contains memes scraped from Reddit. The dataset originally had 1137 samples but only 1060 effective ones after labeling, due to 77 being duplicates.

\item[Indian Memes.] This dataset is a collection of 300 unlabeled memes scraped from ScoopWhoop, an Indian digital media website. The memes are in English but in the Indian context.

\item[jafer covid reddit memes.]  This dataset contains memes about COVID-19 scraped from Reddit. The dataset originally had 671 samples but only 669 effective ones after labeling, due to 2 being duplicates.

\item[MIND-Lab Misogynistic Memes.] This dataset consists of memes that are labeled by domain experts (DE) and crowdsourcing (CS) on whether they are misogynistic, containing aggressiveness, or containing irony. Samples labeled with "misogynisticDE", "aggressiveDE" or "ironyDE" are considered to be offensive and we also take note of the relevant method of attack as misogyny, aggression or irony respectively. As the entire dataset is about misogynistic memes, the victim group is women for the offensive samples. The dataset originally had 800 samples but only 796 effective ones after labeling due to 4 corrupt files. This dataset restricts its usage to research and academic uses only.

\item[memes classified and labelled.] This dataset contains memes scraped from Reddit in 2018. It does not have labels that fit our task. The dataset originally had 5716 samples but only 5685 effective ones after labeling due to 10 corrupt files and 21 rejected by OpenAI content policy.

\item[MemeCap Dataset.] This dataset is a collection of memes from Reddit's r/memes community. They are enriched with annotations, including literal image captions and meme captions with associated visual metaphors. The authors have manually filtered all memes to remove any offensive content, thus we label all samples from this dataset as not harmful. Other labels are not used as they do not sufficiently align with our task's needs. The dataset originally had 6416 samples but only 6375 effective ones after labeling, due to 35 missing labels, 5 corrupt files, and 1 rejected by OpenAI content policy.

\item[MET-Meme.] This dataset contains a mix of Chinese and English memes rich in metaphorical features. The dataset originally had 10039\footnote{The author reported 10045 samples but only 10039 images were provided.} samples but only 10021 effective ones after labeling due to 17 corrupt files and 1 rejected by OpenAI content policy. We use its "offensiveness detection" and "intention detection" labels.

\item[Multi-OFF.] This dataset is a collection of 743 memes related to the 2016 US election. We used its offensiveness label.

\item[Multimedia Automatic Misogyny Identification (MAMI).] The MAMI dataset originally had 11100\footnote{The author's repository's README stated 11000 but the actual sample count was 11100.} samples but only 11081 effective ones after labeling due to 18 corrupt files and 1 rejected by OpenAI content policy. It provides labels for not only whether memes are misogynous, but also what types of misogyny it demonstrates (in the categories of shaming, stereotyping, objectification, and violence). In our dataset, we label the offensive samples' victim groups as women and include the categories identified in the methods of attack field.

\item[r/memes dataset.] This dataset is a collection of 7053 memes posted on Reddit. It does not have labels that fit our task.

\item[Reddit Memes Dataset.] This dataset is a collection of 3326 memes posted on Reddit. It has 3325 effective samples as 1 was rejected by OpenAI content policy. It does not have labels that fit our task.

\item[shinde memes images ocr data.] This dataset contains 6202 memes on COVID-19 and US politics, but the majority of the dataset images were not released, and only 16 were available. The dataset has human-labeled OCR text, victim group, method of attack, and harmfulness classification by us, and GPT-4V generated description and reasoning for our labels.

\item[tamil troll.] This dataset contains memes in Tamil language, a rare resource of one of the four official languages of Singapore. The dataset was originally meant for the classification of "troll" and "not troll" memes. The authors defined "troll" as "a person who upsets or starts a hatred towards people or community", which aligns with our definition of harmful. However, given that the dataset was collected from personal social media and instant messages of Indians, their social standards for a harmful meme can differ from that of Singaporeans. Thus, we did not pass the labels to GPT-4V. The dataset originally had 2967 samples but only 2664 effective ones after labeling due to 3 corrupt files.  This dataset was released under GPLv3 license.

\item[thakkinapalli memes classification.] This dataset has two classes, "meme" and "not meme". We ignored all the images labeled as "not meme". This leaves 753 effective samples after labeling.

\end{description}

\subsection{Singapore-specific meme datasets}
To provide the Singapore context and up-to-date memes, we scraped internet platforms such as the r/Singapore subreddit, Facebook and Instagram accounts posting various themes for memes. \autoref{tab:Singapore-specific meme accounts} summarizes these accounts. The scraped r/Singapore dataset contains memes that were posted with the "Meme" and "SHITPOST" flairs. As there were no labels for scraped data, all labels were fully generated by GPT-4V. We carefully chose these popular-among-locals sources focusing on different themes to ensure that memes collected from them boasts high diversity while being localized and aligned with modern societal interests.
\begin{table}[H]
    \centering
    \caption{Singapore-specific meme Instagram accounts}
    \fontsize{11}{\baselineskip}
        \begin{tabular}{lc}
            \toprule
            \textbf{Account} & \textbf{Theme of content} \\
            \midrule
            @bukittimahpoly & Education \\
            @childrenholdingguns & National Service\tablefootnote{National Service (NS) in Singapore refers to the system where all male Singapore citizens and permanent residents are required by law to serve in the country's army, police or civil defence force for 22 to 24 months.} \\
            @diaozuihotline & General \\
            @memedefsg & National Service \\
            @rafflesplacemrt & Education\\
            @sgagsg & General \\
            @socialstudies.textbook & Education \\
            @socialstudies\textunderscore workbook & Education  \\
            @tkk.jc & Education  \\
            @yourgirlfriendiswhosia & Relationship \\
            A Better World By Memes & Education  \\
            \bottomrule
        \end{tabular}
    \label{tab:Singapore-specific meme accounts}
\end{table}

\begin{table}[H]
    \centering
    \caption{Singapore-specific meme Instagram account themes}
    \fontsize{11}{\baselineskip}
        \begin{tabular}{lc}
            \toprule
            \textbf{Theme} & \textbf{Sample size} \\
            \midrule
            Education & 4938 \\
            General & 19654 \\
            National Service & 2203 \\
            Relationship & 740 \\
            \bottomrule
        \end{tabular}
    \label{tab:Singapore-specific meme themes}
\end{table}

\begin{table*}[t]
    \centering
    \caption{Examples of offensive Singapore context memes}
    \fontsize{11}{\baselineskip}
    \begin{tabular}{p{3cm}p{3cm}p{3cm}p{3cm}p{3cm}}
        \textbf{Meme} &
        \includegraphics[width=3cm]{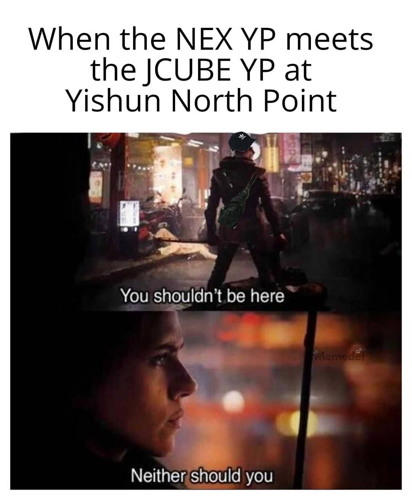} &
        \includegraphics[width=3cm]{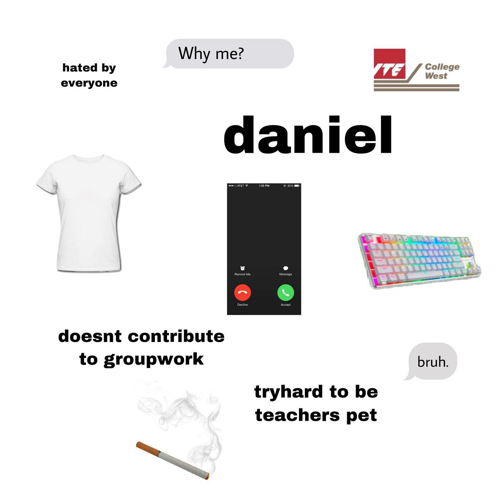} &
        \includegraphics[width=3cm]{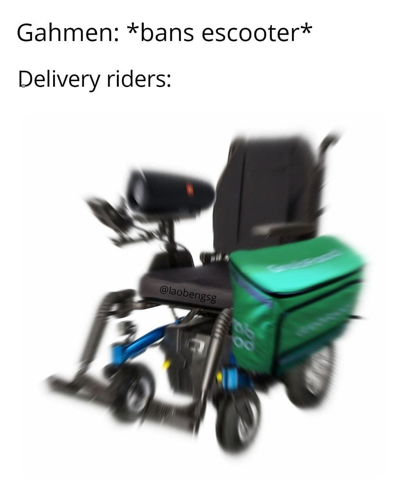} &
        \includegraphics[width=3cm]{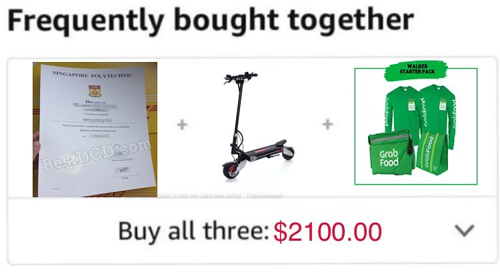} \\
        \hline
        \textbf{Dataset} &
        @diaozuihotline &
        @diaozuihotline &
        @memedefsg &
        @dover\_poly \\
        \hline
        \textbf{Explanation} &
        {\scriptsize Stereotyping that the "Serangoon" and "Jurong East" regions in Singapore (where "NEX" and "JCUBE" shopping malls are located) have many Chinese gangsters. In Singapore slang, "YP" often refers to gangsters of Chinese race.} &
        {\scriptsize Stereotyping that people named "Daniel" are subpar, while portraying students from the ITE West College are people who were disliked, game addicts, smoker and lazy. ITE is an education pathway in Singapore that is less selective, thus some people deem it less elite.} &
        {\scriptsize Mocking that some food delivery workers who are more elderly use electrical wheel chairs to "legally" move on pedestrian pavements (as bikes/scooters are banned on pavements in Singapore). Senior citizens can apply for licenses to ride these vehicles on pavements but some people think that they do not actually need to, and are merely taking advantage of their status to perform deliveries more quickly. In reality, most of them are forced to work due to limited income and high cost of living.} &
        {\scriptsize Mocking that graduates from Singapore Polytechnic (a technical school in Singapore) will ultimately end up as a food delivery man, a job commonly seen by Singaporeans as low-wage and low-skilled. Reflects the prejudices and elitist mindset that many Singaporeans uphold regarding education, believing that the Junior College path (equivalent to a Senior High School, requires the highest academic achievement in Singapore among many education paths) is the most elite. } \\
        \hline
        \textbf{Theme} &
        {\scriptsize Stereotyping based on region, Racism} &
        {\scriptsize Personal attack on name, Stereotyping based on school} &
        {\scriptsize Mockery of elder citizens} &
        {\scriptsize Prejudices on education pathway} \\
        \hline
    \end{tabular}
    \label{tab:Singapore meme example}
\end{table*}

\autoref{tab:Singapore meme example} shows examples of offensive memes that must be understood together with knowledge on Singapore local context. They involve Singapore-specific abbreviations, slang that are not part of any official language, recent changes in Singapore's laws and regulations, typical stereotypes and mindset of Singaporeans etc. We provide explanations in Singapore context to help readers in understanding these examples.

\subsection{Singapore-specific abbreviations}
\label{sec:Singapore-specific abbreviations}
Memes are a highly information-dense medium of communication and abbreviations are often used to reduce the amount of text in the image. We used the Wikipedia article "List of Singapore abbreviations"\cite{sgabbreviations} for a list of abbreviations commonly used in Singapore, constructing a dictionary mapping. In both training and inference, we replace any instance of abbreviations within this dictionary with their full form.

\subsection{Multimodal Singapore Wikipedia Dataset}
To help the model learn more localized, up-to-date knowledge, we scraped Wikipedia articles and images in them by following links 1 level deep from the articles "\{2020, 2021, 2022, 2023\} in Singapore"\footnote{As listed in \url{https://w.wiki/9ytc}}. To construct question-answer pairs that can be used for instruction fine-tuning, we pick from 8 random question templates and set the answer as the text of the article. The question templates are as follows:
\\
\begin{enumerate}[leftmargin=0.8cm]
    \item What is \{title\}?
    \item Explain \{title\} in detail.
    \item Can you explain \{title\} to me?
    \item What exactly does \{title\} entail?
    \item Could you provide some insight into \{title\}?
    \item I'm curious about \{title\}, could you shed some light on it?
    \item Could you elaborate on \{title\} for me?
    \item What's the story behind \{title\}?
\end{enumerate}
\vspace{\baselineskip}
\par
Most articles come with images, and some come with a cover image. In addition to the above text-only question-answer pairs, we construct multimodal question-answer pairs using these images, with one image per pair, and one pair per Wikipedia entry, and the article itself as the answer. We only use cover images as they often contain the most relevant or highest-quality image. We also randomly pick from 9 question templates and place the image in front of texts, following the rest of the text. We also make use of all non-cover images and their alt-texts. Alt-texts are set for readers with disabilities and used by accessibility functions like screen-readers, thus they mostly contain concise descriptions of the image. We filter out images with alt-text shorter than 20 characters as they are often a default value stating the image size, e.g. '150px x 150px'. The multimodal question templates are as follows:
\\
\begin{enumerate}[leftmargin=0.8cm]
    \item What is in this image?
    \item What is the story behind this image?
    \item Can you explain this image to me?
    \item What exactly does this image entail?
    \item Could you provide some insight into this image?
    \item Could you elaborate on this image for me?
    \item Can you give me a detailed rundown of this image?
    \item I'm curious about this image, could you shed some light on it?
    \item Explain this image in detail.
\end{enumerate}
\vspace{\baselineskip}
\par
In summary, we collected 715 instruction following data pairs, including 192 pairs with cover images and detailed descriptions, 157 samples with text only, and 366 pairs of image and alt-text captions. These help the model recognize popular figures and objects in Singapore, some of which are often used in memes. Some examples are: pictures of various political party speakers, logos of local brands and corporations, and public transport vehicles.

\section{Methodology}
\begin{figure}[!h]
  \centering
  \includegraphics[width=\linewidth]{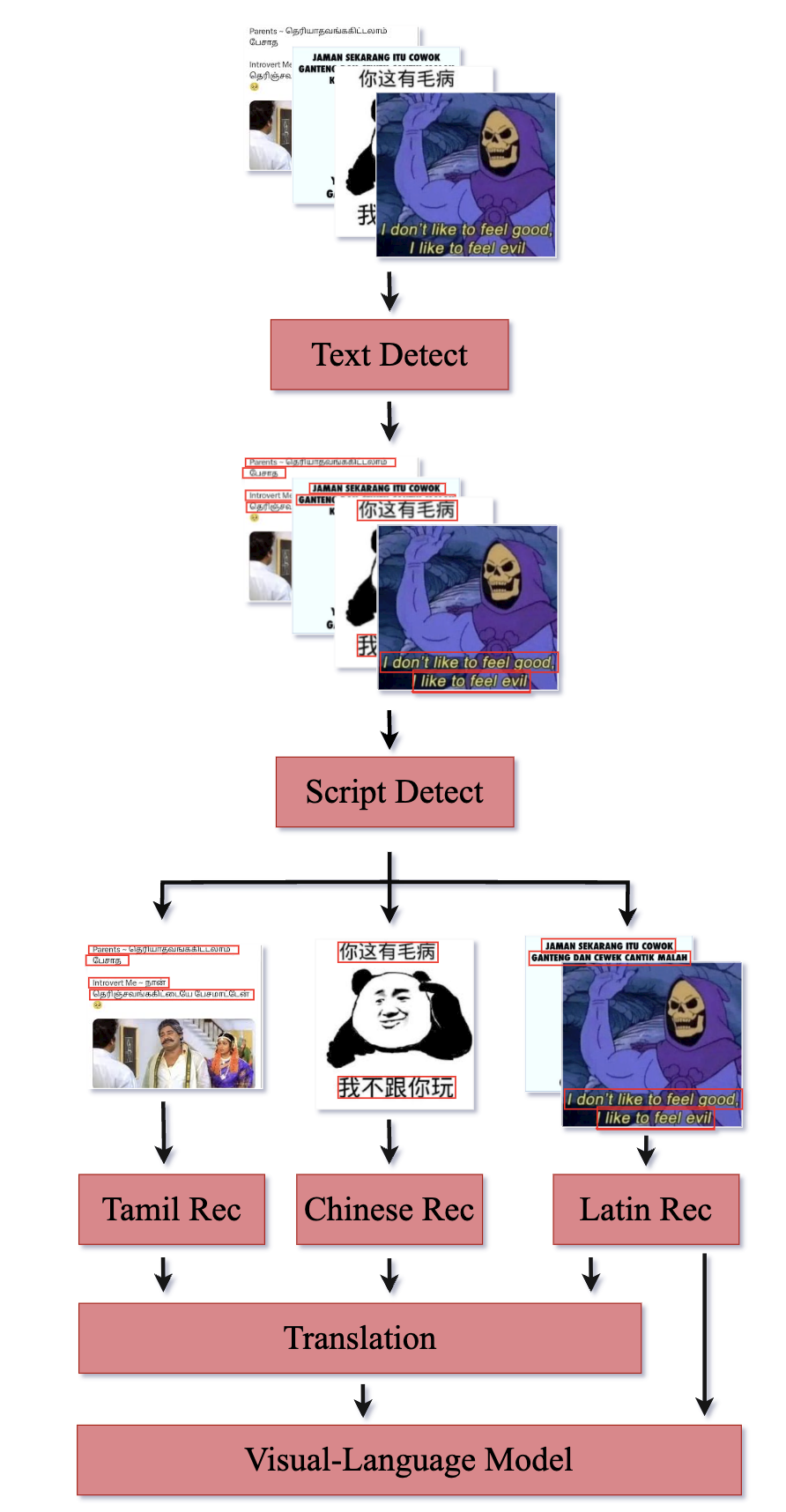}
  \caption{Pipeline}
  \Description{Pipeline}
\end{figure}

\subsection{OCR}
We employ optical character recognition (OCR) as a first step to improve the model's ability to read the text in the sample. This is achieved using the PaddleOCR library, which offers fast multilingual character detection. A first pass is done with the multilingual detection model PP-OCRv4 (successor of PP-OCRv3\cite{ppocrv3}) to detect characters. The bounding boxes are passed to PaddleClas's PPLCNet\_x1\_0\cite{pplcnet}, which detects the language script (Latin, Chinese, or Tamil). In case of multiple scripts detected, the majority will be chosen. If the detected script is not one of the three, it is treated as English and passed directly to VLM. We estimate that the frequency of this happening is extremely low and does not warrant additional compute resources for additional processing. The appropriate language-specific OCR model is then used to provide the final results, to ensure highest accuracy. If any text is detected, abbreviation expansion will be done as mentioned in \autoref{sec:Singapore-specific abbreviations}.

\subsection{Translation}
As the majority of training data used in open-source LLMs today are English-based, LLMs tend to perform best in English. Thus, we translate the text output from the OCR step to English. As the OCR step only classifies between scripts but not languages, we cannot differentiate between languages using the same script, for example, English and Malay. Thus, for Latin scripts, we further detect the language using the langdetect Python package\footnote{As the package does not support Bahasa Melayu, we classify the language as Bahasa Melayu when Bahasa Indonesia is detected since they are highly similar in nature.}, then translate the text to English using Meta's SeamlessM4Tv2 Large\cite{seamlessm4tv2}.

\subsection{Vision-Language Model}
We used LLaVA-NeXT (a.k.a. LLaVA-v1.6) Mistral 7B ("LLaVA") and performed standard LoRA fine-tuning on our dataset. This establishes a baseline for our dataset and evaluation metrics. We then performed the same on Qwen2-VL-7B-Instruct ("Qwen2-VL"). This explores the importance of the base model's performance and pre-trained knowledge on this task since the language model of LLaVA, the Mistral 7B v0.2, is older than Qwen2-VL's Qwen2-7B, and that Qwen2-VL is an overall stronger VLM than LLaVA.
\\
\par
\label{loratricks}
We extend our baseline performance by experimenting with various LoRA-related techniques, including LoRA+\cite{lora_plus}, DoRA\cite{dora}, Rank-Stabalized LoRA\cite{rslora}, and PiSSA\cite{pissa}, on the Qwen2-VL.

\subsection{Prompting}
\label{sec:Prompting}
The output of the OCR and translation model is embedded in the following prompt, following the model's respective prompt template.

\begin{spverbatim}
<image>\n  You are a professional content moderator. Analyze this meme in the context of Singapore society. {ocr_prompt}\n\nOutput a YAML in English using tab for indentation that contains description, the victim groups and methods of attack if any. Think through the information you just provided and label the meme as harmful using "Yes" or "No". Do not include any other explanation outside the YAML.
\end{spverbatim}

\vspace{\baselineskip}
\par
The "\{ocr\_prompt\}" is only present if our pipeline detects text in the picture. If the detected language is not English, it begins with "The text in this meme translated from \{ocr\_lang\} is:" , else, it will begin with "The text in this meme is:", followed by "\textbackslash{}n\{ocr\_text\}".
\\
\par
YAML is empirically a more reliable and token-saving alternative than JSON\cite{yamlvsjson}, especially in our task where the model may output quotation marks, causing JSON decoding to fail. Thus we chose YAML as the output format.

\subsection{Training}
Training of LLaVA was done using 7 NVIDIA RTX 3090 GPUs in 11 hours. Training of Qwen2-VL was done using 4 NVIDIA Tesla V100 32GB PCIE GPUs in 5 hours. Training details can be found in \autoref{sec:TrainingDetails}. Training of both LLaVA and Qwen2-VL followed the recommended hyperparameters of LLaVA since Qwen2-VL did not recommend any for LoRA, with an extended context length of 6144 tokens to fit the longer Wikipedia training samples and allow for a larger room for reasoning.

\subsection{Evaluation}
We evaluated our methods on the above-mentioned validation set consisting 2897 Singapore-context memes. We could not evaluate on any existing global-context meme datasets as they have all been used for training.
\\
\par
We used two metrics to evaluate our solutions: Accuracy and Area Under Receiver Operating Characteristic curve (AUROC)\cite{AUROC}, with the latter as the primary goal since it is more suitable for the unbalanced nature of ground-truths in both our dataset and real-world usage. To locate the correct classification token and its logits, we search for the last token that after stripping white spaces and lowering case, decodes to "yes" and "no". For example, "Yes", "No", "\_yes" and "\_no" where "\_" can be a white space or a preceding byte.
\\
\par
The output score or probability of a harmful meme is calculated by taking the sum of logits of all tokens belonging to the output token's class (since each class can have multiple corresponding tokens after stripping and lowering case), and divide by the sum of logits of all tokens from both classes.

\begin{equation}
    \frac{\sum \text{logits of all tokens of output class}}{\sum \text{logits of all valid output tokens}}
\end{equation}
\\
\par
We did not use constrained decoding techniques as they can distort the logits and thus metrics like AUROC. It also does not accurately reflect the ability of the VLM outputting in the specified format. However, in production, it is still recommended to use them to improve reliability of the system.

\section{Results}
\subsection{Evaluation Results}
\begin{table}[H]
    \caption{LLaVA-NeXT Standalone}
    \resizebox{\linewidth}{!}{
        \begin{tabular}{l|cc}
            \toprule
            \textbf{Model} & \textbf{Accuracy} & \textbf{AUROC} \\
            \midrule
            LLaVA-NeXT-Mistral-7B pre-trained & 0.3316 & 0.5606 \\
            LLaVA-NeXT-Mistral-7B pre-trained w/ sampling & 0.2727 & 0.5543 \\
            LLaVA-NeXT-Mistral-7B (all) & \textbf{0.7259} & \textbf{0.7345}  \\
            LLaVA-NeXT-Mistral-7B (all) w/ sampling & 0.6842 & 0.6868  \\
            \bottomrule
        \end{tabular}
    }
    \label{tab:LLaVA-NeXT Standalone}
\end{table}

\begin{table}[H]
    \caption{LLaVA-NeXT Pipeline}
    \resizebox{\linewidth}{!}{
        \begin{tabular}{l|cc}
            \toprule
            \textbf{Model} & \textbf{Accuracy} & \textbf{AUROC} \\
            \midrule
            LLaVA-NeXT-Mistral-7B pre-trained & 0.4208 & 0.6514 \\
            LLaVA-NeXT-Mistral-7B pre-trained w/ sampling on retry & 0.6500 & 0.6578 \\
            LLaVA-NeXT-Mistral-7B (all) & 0.8064 & 0.7979  \\
            LLaVA-NeXT-Mistral-7B (all) w/ sampling on retry & \textbf{0.8064} & \textbf{0.7991}  \\
            \bottomrule
        \end{tabular}
    }
    \label{tab:LLaVA-NeXT Pipeline}
\end{table}

\begin{table}[H]
    \caption{Qwen2-VL Standalone}
    \resizebox{\linewidth}{!}{
        \begin{tabular}{l|cc}
            \toprule
            \textbf{Model} & \textbf{Accuracy} & \textbf{AUROC} \\
            \midrule
            Qwen2-VL-7B-Instruct pre-trained & 0.6966 & 0.6376 \\
            Qwen2-VL-7B-Instruct pre-trained w/ sampling & 0.6686 & 0.6177 \\
            Qwen2-VL-7B-Instruct (without Wikipedia \& SG memes) & 0.7919 & 0.7575 \\
            Qwen2-VL-7B-Instruct (without Wikipedia) & 0.7967 & 0.7724 \\
            Qwen2-VL-7B-Instruct (all) & \textbf{0.8039} & \textbf{0.7866} \\
            Qwen2-VL-7B-Instruct (all) w/ sampling & 0.7988 & 0.7151 \\
            \midrule
            Qwen2-VL-7B-Instruct DoRA (all) & 0.7960 & 0.7890 \\
            Qwen2-VL-7B-Instruct LoRA+ (all) & \textbf{0.8064} & 0.7991 \\
            Qwen2-VL-7B-Instruct PiSSA (all) & 0.7977 & 0.7998 \\
            Qwen2-VL-7B-Instruct rsLoRA (all) & 0.8043 & \textbf{0.8192} \\
            \bottomrule
        \end{tabular}
    }
    \label{tab:Qwen2-VL Standalone}
\end{table}

\begin{table}[H]
    \caption{Qwen2-VL Pipeline}
    \resizebox{\linewidth}{!}{
        \begin{tabular}{l|cc}
            \toprule
            \textbf{Model} & \textbf{Accuracy} & \textbf{AUROC} \\
            \midrule
            Qwen2-VL-7B-Instruct pre-trained & 0.6969 & 0.6440 \\
            Qwen2-VL-7B-Instruct (without Wikipedia \& SG memes) & 0.7950 & 0.7752 \\
            Qwen2-VL-7B-Instruct (without Wikipedia) & 0.7912 & 0.7747 \\
            Qwen2-VL-7B-Instruct (all) & \textbf{0.8015} & \textbf{0.7862} \\
            \midrule
            Qwen2-VL-7B-Instruct DoRA (all) & 0.8029 & 0.7757 \\
            Qwen2-VL-7B-Instruct LoRA+ (all) & \textbf{0.8036} & 0.8093 \\
            Qwen2-VL-7B-Instruct PiSSA (all) & 0.7943 & 0.8025 \\
            Qwen2-VL-7B-Instruct rsLoRA (all) & 0.7970 & \textbf{0.8130} \\
            \bottomrule
        \end{tabular}
    }
    \label{tab:Qwen2-VL Pipeline}
\end{table}

\begin{table}[H]
    \caption{Qwen2-VL Pipeline with sampling on retry}
    \resizebox{\linewidth}{!}{
        \begin{tabular}{l|cc}
            \toprule
            \textbf{Model} & \textbf{Accuracy} & \textbf{AUROC} \\
            \midrule
            Qwen2-VL-7B-Instruct pre-trained & 0.6962 & 0.6437 \\
            Qwen2-VL-7B-Instruct (without Wikipedia \& SG memes) & 0.7929 & 0.7469 \\
            Qwen2-VL-7B-Instruct (without Wikipedia) & 0.7953 & 0.7929 \\
            Qwen2-VL-7B-Instruct (all) & \textbf{0.8022} & \textbf{0.7868} \\
            \midrule
            Qwen2-VL-7B-Instruct DoRA (all) & 0.8029 & 0.7738 \\
            Qwen2-VL-7B-Instruct LoRA+ (all) & \textbf{0.8036} & 0.8097 \\
            Qwen2-VL-7B-Instruct PiSSA (all) & 0.7922 & 0.7997 \\
            Qwen2-VL-7B-Instruct rsLoRA (all) & 0.7970 & \textbf{0.8133} \\
            \bottomrule
        \end{tabular}
    }
    \label{tab:Qwen2-VL Pipeline with sampling on retry}
\end{table}

\par
We evaluated our pipeline, which consists of OCR, translation to English, and VLM. We also evaluated the VLMs on their own to investigate the usefulness of the supplementary steps in the pipeline, and to accurately reflect the performance of each VLM. In standalone evaluations, OCR results are omitted from the prompt template in \autoref{sec:Prompting}.
\\
\par
We trained the VLMs on 3 variation of training data: "all" refers to combined dataset containing Singapore-related Wikipedia corpus, Singapore-context memes, and generic memes; "without Wikipedia" refers to the latter two combined, and "without Wikipedia \& SG memes" refers to dataset with only generic memes.

\subsection{Standalone Results Analysis}
We find that including the Singapore-context memes significantly increases model accuracy, while including the Singapore Wikipedia dataset contributed less, likely due to the limited tokens in it and the significantly overlapping knowledge with existing LLMs. Training on all data yielded the best results consistently.
\\
\par
Despite having a modest performance pre-trained, LLaVA improves greatly upon fine-tuning, showing that our dataset is effective and contains many knowledge previously unknown to the pre-trained LLaVA model. The stronger performance of pre-trained Qwen2-VL shows that it already has significant foundational knowledge required by the task, possibly because its pre-training focuses on bilingual (English and Chinese) corpus, also the most used languages in Singapore.
\\
\par
Comparing both, results show that pre-trained knowledge plays a significant role even in this domain-specific task and can ultimately determine the upper limit of fine-tuning. That being said, with a rich mix of task-specific training data, the gap between both models significantly closes, highlighting the importance of both pre-trained knowledge and fine-tuning on such a specific and localized task.
\\
\par
Using any of the 4 LoRA related techniques (\autoref{loratricks}) on Qwen2-VL improves the model's performance, with rsLoRA being the best on the AUROC metric, and LoRA+ being the best on accuracy.

\subsection{Pipeline Results Analysis}
LLaVA-NeXT with pipeline showed a great improvement over standalone. This trend continues after fine-tuning, allowing it to match the best Qwen2-VL fine-tuned models. This highlights both the effectiveness of our augmentations to VLM (OCR and translation), and LLaVA's possibly lacking in OCR and multilingual understanding. This also shows OCR and multilingual capability is a crucial part in moderating memes.
\\
\par
Pipeline augmentations did not improve meaningfully Qwen2-VL except for pre-trained and trained without Wikipedia and SG memes. We hypothesize that this could be due to multiple factors.
\\
\par
Firstly, we note that the OpenAI GPT-4V model used to label the dataset is rather dated and is one of the first few versions of frontier VLMs. This means that the label quality might not be high enough for a new model like Qwen2-VL to learn effectively, and there could even be cases where Qwen2-VL model correct and the label is wrong, due to the blur boundary and highly subjective nature of this task.
\\
\par
Secondly, the fact that LLaVA-NeXT with pipeline is able to match Qwen2-VL standalone implies that the language capability of a pre-trained LLM is not as important as the vision understanding capability (e.g. vision encoders and vision feature alignment) for this task. The OCR and translation steps in pipeline makes up for the better vision perception and alignment that Qwen2-VL boasts, which allows it to have strong OCR and multilingual text understanding from complex images without specialized models' help. This reflects the Qwen team's focus on text recognition and multilingual capabilities \cite{qwen2vl}. 
\\
\par
Thirdly, there is inherently an upper limit on how much knowledge a small LLM with 7 billion parameters can learn, especially with LoRA.  We were unfortunately unable to verify this further on larger models or with full-parameter fine-tuning due to our limited resources. It is possible that both VLMs hit a similar capacity bottleneck in its LM due to the smaller parameter count. The vision encoder of Qwen2-VL may also be powerful enough to cause diminishing returns as we augment the pipeline more.
\\
\par
Based on our results, our recommendation is to run Qwen2-VL standalone for maximum accuracy and minimum resource usage.

\subsection{Sampling}
We tested the VLMs and pipeline with sampling (using min-p=0.1 and temperature=0.9). Under pipeline setting, we default to greedy decoding, but if the model fails to output a valid class token, we retry with sampling. In standalone, we apply sampling for all test cases.
\\
\par
Sampling improved LLaVA-NeXT's accuracy when running with pipeline, but reduced it when running standalone. The improvement to trained model in pipeline is negligible too. This shows that sampling does not help the LLaVA-NeXT to get more accurate responses, but helps with cases that results in failed inference like long repetition of gibberish tokens, or wrong output format. However, such issues only happen frequently in pre-trained model, and not so much in fine-tuned model, thus the negligible improvement for the latter.
\\
\par
For Qwen2-VL standalone, sampling greatly worsens performance on the AUROC metric, due to distorted logits. Contrary to our expectations, it also hurts accuracy on both the pre-trained and fine-tuned Qwen2-VL models. Sampling on retry has no meaningful effect on pipelines using Qwen2-VL, likely because Qwen2-VL is stronger at following instructions, thus the chance of a wrongly formatted response is much lower, triggering less retries. In the rare occasions that it does fail, retrying may have introduced more uncertainty and margin for error compared to greedy decoding.
\\
\par
Across both models, sampling did not improve, and often hurt the model's accuracy on the task. This shows that the creativity in generation from sampling is not suitable nor helpful in a classification task like this, even with reasoning-based output instead of direct classification output.

\section{Conclusion}

\par
In this paper, we have shown the effectiveness of multimodal LLMs in content moderation, and their ability to classify highly localized and specific content after fine-tuning. We also show that the pre-trained knowledge plays a large part in the fine-tuned performance, and that the knowledge contained in our training dataset are mostly not covered in the pre-training datasets. We also verified and compared the effectiveness of 4 LoRA-related techniques on the task. Finally, we release our full dataset in 3 variants, full training code and model weights for the 2 VLMs.

\subsection{Limitations}

\textbf{Hallucination.} Similar to LLMs, VLMs might generate outputs that are not grounded in facts or input data. This can lead to inaccuracies in the classification outcome, thus reducing effectiveness of the moderation system.
\\ \\
\textbf{Biases.} As our data were mostly labeled with GPT-4V, any biases or flaws in the model will directly impact the labeled data, thus any fine-tuned models on it. However, we expect the strong content filtering mechanism and alignment efforts by OpenAI to minimize the chance of biased labels from GPT-4V. Biases can also be transferred from the base models of our fine-tuned models, both from the vision encoder and the language decoder. This may lead to biased outcomes or unfair representations of certain contents, especially given that some offensive memes are already of such nature, potentially causing false negatives (e.g. allowing an offending meme to pass). In these cases, existing biases may be amplified as the end user consumes such content.
\\ \\
\textbf{Label quality} Newer and better performing models have been released compared to the model we used to label the dataset. This means the dataset and its labels might not be in the most accurate form as one can get today, using a latest model as labeler. Any trained model on this dataset will therefore be limited to the accuracy of the labeling model we used, which as time passes, may become worse than a pre-trained model. In this case, fine-tuning a pre-trained model would lead to worse result.

\subsection{Potential Misuse}
Our work may be misused in serval ways. One can use our model adversarially to train content generating models, or reverse engineer contents that are both offensive but able to pass our moderation. One can also use the model to curate a dataset containing offensive content only, and further use it to train malicious models that can promote societal biases and hate online. To prevent such from happening, we will release the dataset in a gated manner, requiring each user to sign an agreement that prohibits them from such uses beyond academic research. Publishing our research will also help the community to understand the issue better and build better moderation mechanisms to filter out offensive content, should they be generated.

\subsection{Future Work}
The 7-billion parameter-class VLMs that we explored already run considerably fast on modern serving hardware (e.g. GPUs) after quantization and deployment optimizations (e.g. TensorRT-LLM). However, explorations into recent, smaller VLMs could be more valuable for large-scale production usage, as we discovered that the relatively small VLMs we tested were able to learn and converge well on our dataset. More investigation can also be done on full-parameter fine-tuning of models instead of LoRA-based fine-tuning only.
\\
\par
Human expert-labeled test set could also prove valuable to evaluate our work with greater accuracy and reliability, as the current evaluation depends on the underlying MLLM used to label the dataset. 

\section{Acknowledgments}
Jiayang, Yuxuan, Sherman and Alistair contributed to the collection and processing of meme datasets. Theodore contributed to Wikipedia and Instagram scraping. Jiayang led training efforts for LLaVA models while Yuxuan led training efforts of Qwen2-VL models and evaluation of both VLMs. Jiayang, Yuxuan and Sherman contributed to the inference pipeline and its performance optimizations. Sherman and Bryan provided training-related advisories and compute resources. Yuxuan, Bryan and Alistair contributed to writing the paper.
\\
\par
We would like to thank the Nanyang Technological University High Performance Computing Club and the National Supercomputing Centre Singapore for their compute resources. Bryan S. would also like to thank Au B. for their help and support.
\\
\par
This research started as an entry to the \href{https://ospc.aisingapore.org/}{\color{blue}Online Safety Prize Challenge} hosted by AI Singapore from February to April 2024. It was then completed during July to October 2024. All views and results presented in this paper do not reflect those of AI Singapore, nor has AI Singapore endorsed or supported this research in any way other than providing the public definition of offensive memes.

\clearpage

\bibliographystyle{ACM-Reference-Format}
\bibliography{biblography}

\clearpage

\appendix

\section{VLM Training Details}
\label{sec:TrainingDetails}

\subsection{LLaVA-NeXT}
\begin{table}[H]
    \centering
    \caption{LLaVA-NeXT Software \& Hardware}
    \resizebox{\linewidth}{!}{
        \begin{tabular}{lc}
            \toprule
            OS & Ubuntu 22.04.4 LTS w/ Linux 5.15.0-122-generic \\
            NVIDIA Driver &  550.54.15 \\
            Python & 3.10.12 \\
            CUDA & 12.4 \\
            PyTorch & 2.4.1+cu124 \\
            Framework & haotian-liu/LLaVA modified \\
            Distributed & DeepSpeed ZeRO-2 \\
            GPU & 7x NVIDIA GeForce RTX 3090 \\
        \bottomrule
        \end{tabular}
    }
    \label{tab:LLaVA-NeXT Software & Hardware}
\end{table}

\begin{table}[H]
    \centering
    \caption{LLaVA-NeXT Hyperparameters}
    \resizebox{\linewidth}{!}{
        \begin{tabular}{lc}
            \toprule
            Context length & 6144 \\
            All seeds & 42 \\
            LoRA Rank & 128 \\
            LoRA Alpha & 256 \\
            LoRA Dropout & 0.0 \\
            Per GPU training batch size & 2 \\
            Gradient accumulation steps\tablefootnote{\label{gacc footnote}Due to a known \href{https://github.com/huggingface/trl/issues/2175}{\color{blue}bug} in the version of Hugging Face transformers library that we used, changing the gradient accumulation steps will not be able to reproduce our results, even if the effective global batch size matches.} & 9 \\
            Global training batch size\tablefootnote{LLaVA officially recommends 128 but due to hardware limitations we had to use 126.} & 126 \\
            Per GPU evaluation batch size\tablefootnote{\label{eval bs footnote}Due to hardware-induced numerical inaccuracies in GEMM, changing the evaluation batch size can change the logits thus AUROC, and sometimes even accuracy.} & pipeline: 1, standalone: 4 \\
            Epochs & 1 \\
            Learning Rate & 1e-4 \\
            Learning Rate Scheduler & Cosine Annealing Decay \\
            LR warm up ratio & 0.03 \\
            Optimizer & AdamW \\
            Precision & BF16 \\
            min\_p (sampling for eval only) & 0.1 \\
            temperature (sampling for eval only) & 0.9 \\
        \bottomrule
        \end{tabular}
    }
    \label{tab:LLaVA-NeXT Hyperparameters}
\end{table}

\subsection{Qwen2-VL}

\begin{table}[H]
    \centering
    \caption{Qwen2-VL Software \& Hardware}
    \resizebox{\linewidth}{!}{
        \begin{tabular}{lc}
            \toprule
            OS & Ubuntu 22.04.4 LTS w/ Linux 6.5.0-27-generic\\
            NVIDIA Driver &  550.54.14 \\
            Python & 3.10.12 \\
            CUDA & 12.4 \\
            PyTorch & 2.4.1+cu124 \\
            Framework & \href{https://github.com/aliencaocao/LLaMA-Factory/tree/aisg-meme}{\color{blue}LLaMA-Factory 0.9.0 modified} \\
            Distributed & DeepSpeed ZeRO-2 \\
            GPU & 4x NVIDIA Tesla V100-32GB PCIE \\
        \bottomrule
        \end{tabular}
    }
    \label{tab:Qwen2-VL Software & Hardware}
\end{table}

\begin{table}[H]
    \centering
    \caption{Qwen2-VL Hyperparameters}
    \resizebox{\linewidth}{!}{
        \begin{tabular}{lc}
            \toprule
            Context length & 6144 \\
            All seeds & 42 \\
            LoRA Rank & 128 \\
            LoRA Alpha & 256 \\
            LoRA Dropout & 0.0 \\
            LoRA+ ratio (for LoRA+ only) & 16 \\
            Per GPU training batch size & 2 \\
            Gradient accumulation steps\tablefootnote{See footnote \autoref{gacc footnote}} & 16 \\
            Global training batch size & 128 \\
            Per GPU evaluation batch size\tablefootnote{See footnote \autoref{eval bs footnote}} & pipeline: 1, standalone: 10 \\
            Epochs & 1 \\
            Learning Rate & 1e-4 \\
            Learning Rate Scheduler & Cosine Annealing Decay \\
            LR warm up ratio & 0.1 \\
            Optimizer & AdamW \\
            Precision & FP16 w/ AMP \\
            min\_p (sampling for eval only) & 0.1 \\
            temperature (sampling for eval only) & 0.9 \\
        \bottomrule
        \end{tabular}
    }
    \label{tab:Qwen2-VL Hyperparameters}
\end{table}

All VLMs are ran at batch size 1 in pipeline evaluation to simulate real-world usage.

\section{Pipeline Details}
\label{sec:PipelineDetails}
\begin{table}[H]
    \centering
    \caption{OCR and translation pipeline}
    \resizebox{\linewidth}{!}{
        \begin{tabular}{lc}
            \toprule
            Python & 3.10.12 \\
            PyTorch & 2.4.1+cu124 \\
            \midrule
            PaddlePaddle-GPU & 2.6.0 \\
            PaddleOCR & 2.7.5 w/ bug fixes \\
            PaddleClas & 2.5.2 \\
            OCR inference batch size & 1024 \\
            OCR inference precision & FP32 \\
            \midrule
            fairseq2 & 0.2.0 @ 76015a1 \\
            fairseq2n & 0.2.0 @ 76015a1 w/ CUDA\_ARCH 8.6 \\
            seamless\_communication & 0.1.0 @ 75ed7ef w/ patches \\
            Translation inference batch size & 24 \\
            Translation inference precision & FP16 \\
        \bottomrule
        \end{tabular}
    }
    \label{tab:OCR and translation pipeline}
\end{table}

\begin{table}[H]
    \centering
    \caption{OCR models}
    \resizebox{\linewidth}{!}{
        \begin{tabular}{lccc}
            \toprule
            Language & Detect & Lang. Classification & Recognition \\
            \midrule
            English & \href{https://paddleocr.bj.bcebos.com/PP-OCRv3/english/en_PP-OCRv3_det_infer.tar}{\color{blue}en\_PP-OCRv3} & \href{https://paddleclas.bj.bcebos.com/models/PULC/inference/language_classification_infer.tar}{\color{blue}PPLCNet\_x1\_0} & \href{https://paddleocr.bj.bcebos.com/PP-OCRv4/english/en_PP-OCRv4_rec_infer.tar}{\color{blue}en\_PP-OCRv4} \\
            Chinese & \href{https://paddleocr.bj.bcebos.com/PP-OCRv4/chinese/ch_PP-OCRv4_det_infer.tar}{\color{blue}ch\_PP-OCRv4} & \href{https://paddleclas.bj.bcebos.com/models/PULC/inference/language_classification_infer.tar}{\color{blue}PPLCNet\_x1\_0} & \href{https://paddleocr.bj.bcebos.com/PP-OCRv4/chinese/ch_PP-OCRv4_rec_infer.tar}{\color{blue}ch\_PP-OCRv4} \\
            Latin (Malay) & \href{https://paddleocr.bj.bcebos.com/PP-OCRv3/multilingual/Multilingual_PP-OCRv3_det_infer.tar}{\color{blue}Multilingual\_PP-OCRv3} & \href{https://paddleclas.bj.bcebos.com/models/PULC/inference/language_classification_infer.tar}{\color{blue}PPLCNet\_x1\_0} & \href{https://paddleocr.bj.bcebos.com/PP-OCRv3/multilingual/latin_PP-OCRv3_rec_infer.tar}{\color{blue}latin\_PP-OCRv3} \\
            Tamil & \href{https://paddleocr.bj.bcebos.com/PP-OCRv3/multilingual/Multilingual_PP-OCRv3_det_infer.tar}{\color{blue}Multilingual\_PP-OCRv3} & \href{https://paddleclas.bj.bcebos.com/models/PULC/inference/language_classification_infer.tar}{\color{blue}PPLCNet\_x1\_0} & \href{https://paddleocr.bj.bcebos.com/PP-OCRv4/multilingual/ta_PP-OCRv4_rec_infer.tar}{\color{blue}ta\_PP-OCRv4} \\
        \bottomrule
        \end{tabular}
    }
    \label{tab:OCR models}
\end{table}

\autoref{tab:OCR and translation pipeline} shows pipeline software details. \autoref{tab:OCR models} shows list of OCR models used. The translation model used was \href{https://huggingface.co/facebook/seamless-m4t-v2-large}{\color{blue}facebook/seamless-m4t-v2-large}.

\section{License}
Due to our usage of two GNU GPL-series licensed datasets in our combined dataset, we have to release our combined dataset under the GNU GPLv3 license. However, the trained model and the full training code, as well as any dataset processing code (we did not use any from original dataset authors), will be released under the MIT license.

\end{document}